# Topic Modeling on Clinical Social Work Notes for Exploring Social Determinants of Health Factors

Shenghuan Sun[1],
Travis Zack[1,4],
Madhumita Sushil[1†],
Atul J. Butte * [1, 2, 3†]

1. Bakar Computational Health Sciences Institute, University of California, San Francisco, San Francisco, CA, USA
2. Center for Data-driven Insights and Innovation, University of California, Office of the President, Oakland, CA, USA
3. Department of Pediatrics, University of California, San Francisco, CA, 94158, USA
4. Division of Hematology/Oncology, Department of Medicine, UCSF, San Francisco, California, USA.

*Author to whom correspondence should be addressed.
† Equal Contribution

WORD COUNTS

ABSTRACT CURRENT WORD COUNT: 248

MAIN TEXT WORD COUNT: 3872 words excluding the structured abstract, tables, figures, references, and acknowledgments.

# ABSTRACT

## OBJECTIVE
Most research studying social determinants of health (SDoH) has focused on physician notes or structured elements of the Electronic medical record (EMR). We hypothesize that clinical notes from social workers, whose role is to ameliorate social and economic factors, might provide a richer source of data on SDoH. We sought to perform topic modeling to identify robust topics of discussion within a large cohort of social work notes.

## MATERIALS AND METHODS
We retrieved a diverse, deidentified corpus of 0.95 million clinical social work notes from 181,644 patients at the University of California, San Francisco. We used word frequency analysis and Latent Dirichlet Allocation (LDA) topic modeling analysis to characterize this corpus and identify potential topics of discussion.

## RESULTS
Word frequency analysis identified both medical and non-medical terms associated with specific ICD10 chapters. The LDA topic modeling analysis extracted 11 topics related to social determinants of health risk factors including financial status, abuse history, social support, risk of death, mental health. In addition, the topic modeling approach captured the variation between different types of social work notes and across patients with different types of diseases or conditions.

## DISCUSSION
We demonstrated topic modeling as a powerful tool to extract latent topics from clinical notes, serving as an ideal data exploration approach.

## CONCLUSION
Social work notes contain rich, unique, and otherwise unobtainable information on an individual's SDoH. These notes contain robust and coherent topics of discussion that can be identified and utilized to evaluate impact SDoH factors have on patient/public-health.

**OBJECTIVE**

Social determinants of health (SDoH), non-medical conditions that influence health, are a significant contributor to health disparities due to systemic disadvantages and bias[1,2]. A better understanding of the role of SDoH in diabetes, heart disease, and other conditions has led to increased attention for the medical system to address these factors in the context of treating these conditions[3–7]. However, our capacity to research these correlations is still quite constrained. Most assessments of SDoH are not present in structured data that is easily accessible to researchers[8]. Instead, much of this information is collected in unstructured clinical social work notes.

Social work notes, compared to standardized structured electronic medical records data, are more challenging to analyze given their unstructured format. Inability to easily extract this information limits research into the effects of SDoH on care delivery and success. Topic modeling methods based on Latent Dirichlet Allocation (LDA) are among the most popular approaches and have been shown to be able to find hidden structures (topics) in large corpora in an unsupervised manner[9,10].

To understand the information embedded in the social work notes, we applied LDA topic modeling to characterize the specific SDoH factors covered across nearly one million clinical social work notes. To our knowledge, we analyzed one of the biggest social work notes collections so far and these notes spanned a diverse set of patient demographics and diseases. This allowed us to develop a comprehensive understanding of the underlying topics from different types of notes or for a variety of disease chapters. Understanding the limitations of topic modeling including fixed number of clusters, intrinsic randomness, and need for human-based interpretation, we used several evaluation approaches to minimize these potential biases.

**BACKGROUND AND SIGNIFICANCE**

Computational understanding of the free text in clinical notes is well known to be an open challenge, including the extraction of structured information from these documents[11]. Some progress has been made in extracting SDoH factors from clinical text using named entity recognition (NER), an NLP method of extracting pre-defined concepts from text[12,13]. Both machine learning-based and traditional rule-based NER have been developed and tested[13–15]. While NER approaches have been shown to be effective, they can be time-consuming and present challenges with regard to bias. For rule-based models, manual vocabulary tuning requires a significant amount of human effort and is prone to a developer's bias on language relevance and meaning. Machine learning models similarly require manual annotation and additionally need sufficient data on which to train models. Both are often tedious, error-prone, and potentially biased by human experiences and research scope. Moreover, the supervised nature of these approaches endangers them to propagating the biases researchers may hold regarding topic or term importance. In order to minimize the biases and limits of NER tasks in clinical research, rigorous and detailed data exploration is critical and strongly recommended before the use of these manual efforts[16].

Topic modeling methods are a set of powerful techniques that have been widely applied towards unbiased topic discovery from unorganized documents[17–19] and have been used in the fields of social science[20,21], environmental science[21], political science[22], and even in biological and medical contexts. However, to our knowledge, LDA topic modeling has not been heavily used to assess corpora of social work notes for SDoH factors, likely due to the general unavailability of large enough corpora.

Clinical social workers are licensed professionals that specialize in identifying and removing social and environmental barriers an individual patient is experiencing. In particular, clinical notes generated by clinical social workers are an invaluable data resource for understanding SDoH information in patients. As such, the clinical notes written by social workers, often include specific text capturing an individual's SDoH. Yet, to date, social work notes have been a relatively under-utilized data source and have not been extensively investigated for understanding SDoH[23].

To our knowledge, this is the first topic modeling study across clinical social work notes. This work demonstrates how unsupervised topic modeling approaches can elucidate and categorize the information relating to SDoH within unstructured clinical notes. It further highlights the rich source of SDoH information clinical social worker notes represent and develops methods which can make further research within the field of SDoH more tractable.

## MATERIALS AND METHODS

### Study design and cohort selection

This study uses the deidentified clinical notes at UCSF recorded between 2012 and 2021[24]. The study was approved by the Institutional Review Board (IRB) of the University of California, San Francisco (UCSF; IRB #18-25163). We collected clinical notes whose metadata contain the term "social" (case insensitive) in either the encounter type, encounter department name, encounter department specialty, or authentication provider type and collectively refer to these documents as 'social work notes'. In this manner, we obtained a subset of 2.5 million social work notes from a corpus of 106 million notes. Notes shorter than 30 characters were removed because they are likely to contain null values or not be informative. We also removed the duplicated notes to reduce the redundancy and computation time. Finally, 0.95 million notes were retained for the downstream topic modeling analysis (**Figure 1**).

### Word frequency calculation

To investigate the disease-specific features in the social work notes, we first used ICD-10 codes to extract 10 ICD-10 chapters: (1) Diseases of the nervous system (G00-G99), (2) Diseases of the circulatory system (I00-I99), (3) Diseases of the respiratory system (J00-J99), (4) Diseases of the digestive system (K00-K95), (5) Diseases of the musculoskeletal system and connective tissue (M00-M99), (6) Diseases of the genitourinary system (N00-N99), (7) Pregnancy, childbirth and the puerperium (O00-O9A), (8) Congenital malformations, deformations and chromosomal abnormalities (Q00-Q99) (9) Neoplasms (C00-D49), (10) Diseases of the blood and blood-forming organs and certain disorders involving the immune mechanism (D50-D89). Python package *scikit-learn* was used to conduct the analysis[25]. To embed and tokenize the unstructured notes, we used *text.CountVectorizer* function from *sklearn.feature* extraction package. We computed the chi-squared statistics with the chi2 function from *sklearn.feature selection*. We compared each note category against all the other notes. After ranking the P values and removing stop words, the top five potential meaningful words were visualized by the word frequency calculation.

**Latent Dirichlet Allocation (LDA) analysis and topic models**
Latent Dirichlet Allocation (LDA) is a generative probabilistic model[19]. It assumes that each document is a combination of a few different topics, and that each word's presence can be attributed to particular topics in the document. The result is a list of clusters, each of which contains a collection of distinct words. The combination of words in a cluster can be used for topic model interpretation. LDA topic modeling can be considered as a clustering algorithm since it takes a collection of documents as vectors of word counts and clusters the data points into a predefined number of cluster centers, which corresponds to the topics. Python package *gensim* was used for the implementation[26]. We used *gensim.models.ldamodel.LdaModel* for the actual analysis. The core estimation code is based on *Hoffman et al*[27].

Text preprocessing was performed before running the topic modeling algorithms, for which python package *nltk* was implemented. First, the special characters including '\t', '\n', '\s' were excluded. The common stop words were also excluded, using *stopwords.words('english')* from *nltk* package[28].

**Determine the cluster number**
One of the most important parameters that LDA analysis needs is the number of topics K. Generally, if K is chosen to be too small, the model will lack the capacity to provide a holistic summary of complex document collections; and returned topical vectors may combine semantically unrelated words/tokens[29]. Conversely, if K is chosen to be too large, the returned topical vectors may be redundant, and a parsimonious explanation of a complex phenomenon may not be achieved. Instead of using the human-in-the-loop method, we used two evaluation metrics to symmetrically determine the optimal cluster number.

The two metrics that we used are Topic Coherence[30] and Topic Similarity[31]:
(1)    Topic Coherence measures the score of a single topic by measuring the degree of semantic similarity between high-scoring    words in the topic (34). These measurements help distinguish between topics that are semantically interpretable topics and topics that are artifacts of statistical inference. For the coherence metric that we used, the measure is based on a sliding window, one-set segmentation of the top words and an indirect confirmation measure that uses normalized pointwise mutual information (NPMI) and the cosine similarity[32].
(2)    Topic Similarity measures how similar two clusters are considering the words that topics contain. The lower the values are, the less redundant the topic distribution is. For that, we use Jaccard similarity[31].
An ideal solution would have a high topic coherence and low similarity metric. To decide the optimal number of clusters, for each analysis we ran the LDA analysis with number of clusters K from 10 to 50, and the C and S were simultaneously calculated. Number of clusters will be chosen if its C value is the $i_{th}$ highest and S value is the $j_{th}$ smallest and i + j is the minimum among all runs (**S.Figure 1A**).

**Topic labeling heuristics**

Our analysis screened nearly 20 topic clusters for all 14 categories of notes for 5 independent times resulting in more than 1000 topic clusters that required further labeling. To avoid human variation and save time, we developed a heuristic to automatically assign topic labels while minimizing human efforts. The details are shown below:

We first constructed the dictionary of topic names and the corresponding words by looking through the topic modeling results for one run on the complete corpus of 0.95 million social work notes at UCSF. Then we expanded the individual topic clusters by first finding 20 most similar words to the words comprising topic clusters based on the cosine similarity of their word embeddings[33]. Any generic words included thus were removed manually. The final dictionary is shown in **S.Table 3**. We thereby automatically assign the topic name for individual word clusters by computing the ratio of intersection versus union of words in a cluster. In this manner, we were able to find topics for all the 1000+ topic clusters the analysis gave. The details are explained in the pseudo-code below***:

*Heuristics of automatic assigning topic names for the individual topic cluster*
**Begin:**
*Construct the dictionary of topic names and the words comprising this topic;*
*Expand the individual topic cluster space;*
*> Enrichment[k|h] means the frequency of words belong to topic h for word cluster k*
*> ∩ means intersect, ∪ means union*
**For** *each iteration of the topic modeling results* **do:**
   **For** *every word cluster k* **do:**
     **For** *topic name h, corresponding word set* **in** *dictionary* **do:**
       *overlap = word cluster k ∩ topic h*
     *total = word cluster k ∪ topic h*
      *Enrichment[k|h] = overlap / total*
    *The topic assigned to word cluster k = max([Enrichment[k|h] for h in H])*
**End**

Code for the paper is available on *https://github.com/ShenghuanSun/LDA_TM*

**RESULTS**

We retrieved a total of 0.95 million de-identified clinical social work notes generated between 2012 and 2021 (see **Methods**) from our UCSF Information Commons[24] (**Figure 1**). The majority of notes were classified as Progress Notes, Interdisciplinary Notes, or Telephone Encounter Notes; other note categories included Patient Instructions, Group Note, Letter, which comprised fewer than 5 percent each. These notes cover 181,644 patients of which 95387 (52.5%) were female. The median age of these patients was 33 years. Among them, 69,211 patients had only one note; 65,100 patients had between 2 and 5 notes, and 47,333 patients had more than 5 notes (**S. Table 1, Figure 2B**). No demographic feature was statistically associated with the number of notes for each patient (**S. Table 1**).

We next studied which medical conditions were being attributed to these patients receiving social work notes. We collected the ICD-10 codes for the encounters during which social work notes were recorded for the patients. These ICD-10 codes were then mapped at the chapter level[34]. The three most frequent ICD-10 chapters found to be associated with a social work note were "Mental, Behavioral and Neurodevelopmental disorders", "Factors influencing health status and contact with health services", and "Symptoms, signs and abnormal clinical and laboratory findings, not elsewhere classified" (**S. Table 2**).

**Word frequency on individual disease**

In order to uncover disease-specific information in social work notes, we compute chi-squared stats of each word feature for the individual disease chapter versus the other chapters (see **Methods**). After removing frequent words, we calculated the word frequency for the top five words with lowest p-values in social work notes associated with each ICD-10 chapter (**Figure 3**).

We found that the social work notes associated with each ICD-10 chapter have disease-specific terms and even some disease-specific social determinants of health topics. For instance, notes from patients with neoplasms have a significant enrichment for cancer-related topics: "oncology", "chemotherapy", and "tumor". Similarly, notes from patients with musculoskeletal disorders had disease-related words like "arthritis" and "rheumatology". However, there were also general and disease-specific patterns of enrichment in SDoH-related words. For example, 'planning' and 'mother' were present across many disease categories, while 'mindfulness' was enriched in *Pregnancy* and *Nervous system, while* "wheelchair" was enriched in *Musculoskeletal disorders.* Notably, pregnancy-related conditions showed a very significant enrichment for potentially relevant SDoH topics. Some of these notes are clearly related to mental health, demonstrating that mental health might be frequently assessed in social work notes during pregnancy care.

**Using LDA to extract topics in social work notes**

While word frequency calculations can provide a window on term relevance, this view is too limited to understand what broader topics may be contained within these notes. In contrast, topic modeling is a field of unsupervised learning that learns statistical associations between words or groups of words to identify "topics": clusters of words that tend to co-occur within the same document.

We implemented Latent Dirichlet Allocation (LDA), a generalized statistical model that allows topic discovery and semantic mining from unordered documents (see **Methods**)[19]. A required input to the LDA model is the desired number of clusters to partition data into. This requirement to predefined number of data partitions could introduce bias, so we used a combination of statistical evaluations of model results to find the ideal partition number (see **Methods**). This approach resulted in 17 topic clusters (see **S.Figure 1, Methods**).

Looking at the word components of each topic (**Table 1**), we discovered a few diverse clusters that cover many different social aspects of patients including social service (Topic 11), abuse history (Topic 14), phone call/ online communications (Topic 12), living condition/ lifestyle (Topic 16), risk of death (Topic 8), group session (Topic 7), consultation/ appointment (Topic 5), family (Topic 4, 6), and mental health (Topic 1). Many of these topics are consistent with topics covering social determinants of health; most importantly, most of the information potentially conveyed through these topics are absent in the structured data. Of note, in our parameter exploration, we found that increasing the number of clusters can lead to additional recognizable topics, such as food availability (data not shown), although we also obtain redundant topics. For the remainder of our work, we continued with computationally determined topic clusters number.

**Topic modeling on specific note categories**
After modeling these topics across all the social work notes, we discovered that certain topics appear more often than the other topics (**Table 1**). To explore whether this was due to notes having topic subtypes of varying size, we compared the identified topics across the four largest categories of social work notes: Progress Notes, Interdisciplinary, Telephone encounters, and Group Notes (**Figure 2A**, see **Methods**). We focused on the major topics in these notes because 1) the rare categories (less than 5 percent) were less likely to influence the topic modeling results; 2) every hospital system is likely to have notes within these categories and thus the analysis results could be more informative and generalizable. We reused the same pipeline on identifying the optimal number of cluster (see **Methods**). Due to the intrinsic randomness that the LDA method has, we ran each analysis five times per category and then developed a heuristic for labeling these topic clusters based on our previous findings for all the notes (see **Methods**).

We found that social work notes that belong to the Progress Notes category, compared those that belong to all the other categories, were composed of clinically related topics (e.g., *Clinician/Hospital/Medication; Mental Health*), along with a smaller proportion of SDoH-related topics (e.g., *Insurance/Income, Abuse history, Social support*, *Family*). This is consistent with the fact that progress notes are routinely collected medical records where healthcare professionals record information to document and update a patient's clinical status. Compared to these, telephone encounter notes are usually used to address patient issues outside of an appointment. Accordingly, telephone encounter notes contain more information about *Insurance/Income*; *Phone call/Online*; *Social support*; *Family*. Interestingly, telephone encounter notes lack information about the *Risk of death* (**Figure 4A**). While telephone encounters may serve to discuss *Family/Social support*, it's possible that, given the severity of this topic, discussions on *Risk of death* are considered not appropriate for telephone encounters and saved for formal in-person visits. In addition, group notes, which are the notes taken during group therapy, describe the group's progress and dynamics. As expected, group notes have more uneven topic category distributions (**Figure 4A**).

We also applied LDA analysis to the social work notes associated with 10 ICD-10 chapters used earlier (**Figure 4B**). We observed that individual diseases have a similar topic proportion distribution, indicating that social workers have a similar procedure when communicating with patients with different diseases. The majority of the clusters are focused on social support and family. This is similar to the topic distribution for Interdisciplinary notes, which intuitively makes sense given the social work notes for each ICD-10 chapter contain notes in a mixture of different subtypes mentioned above. However, they indeed have some differences: compared with other disease chapters, notes associated with disorders of mental health and pregnancy contain more abundant SDoH topics on mental health, as would be expected. Mental health topics are even more often mentioned in clinical notes around pregnancy than even nervous system disorders. Interestingly, the family topic area was often mentioned in notes associated with birth malformation abnormalities. In summary, this analysis demonstrates both the commonness and uniqueness of topics around social determinants of health covered across the various diseases and conditions which afflict patients.

**DISCUSSION**

We used a comprehensive unsupervised topic modeling method called LDA modeling on our corpus of 0.95 million de-identified clinical social work notes. We showed that topic modeling can (1) extract the hidden themes from this huge corpus of clinical notes and identify the critical information embedded in the notes, namely social determinants of health (SDoH) factors; and (2) calculate the proportion of each theme across the note corpus and systemically characterize notes of different types.

Using simple term frequency methods on this large corpus, we found that specific SDoH terms tend to be enriched in notes from patients within different diseases categories, including wheelchair for patient with musculoskeletal disorders and depression for patients with pregnancy diagnoses, suggesting that these populations may be more at risk for these SDoH features.

We performed LDA modeling on this large corpus, which allowed us to extract topic categories from this large corpus of notes. Through this method, we extracted several SDoH-related topics that are intuitive, provide insight into the information that may be extracted from these corpuses, and can be leveraged in future work to understand how these topics correlate with health outcomes. During our comparison of notes of different subtypes, we found that the distribution of SDoH topics in notes of different categories varies. Interestingly, the topic distribution of notes for specific types of diseases contains similar information but showed different levels of enrichment that represent the unique features of each disease set. As one of many examples, our analysis shows how mental health issues are frequently documented around pregnancy (**Figure 4B**). This type of information can help us better understand the social determinants of most concern to patients and when interacting with the health system.

The specific topics that we identified are consistent with a few previous publications. A recent study implemented non-negative matrix factorization (NMF) topic modeling method over the 382,666 primary care clinical notes and were also able to extract information regarding physical, mental, and social health. However, this study contained clinical notes generated solely by physicians, while we focused specifically on social work notes which allowed a more comprehensive list of SDoH topics to be identified. Our work is a step closer towards understanding SDoH-related information that is embedded in clinical social work notes and we believe that compared with other supervised methods, unsupervised approaches may be better at identifying coherent, yet comprehensive topics.

Our study has several strengths. We implemented a comprehensive topic modeling analysis on a huge corpus of notes that to our knowledge is the largest social work notes data set. Instead of focusing on a single disease category or specific medical topic, we are aiming at comprehensively finding the potential SDoH topics in all types of clinical social notes for a variety of diseases. Moreover, in order to provide an initial, comprehensive landscape on the embedded information in social worker notes, we performed both a word frequency enrichment analysis, which identified specific terms that more frequently appear in conjunction with specific ICD-10 chapters, and unsupervised topic modeling approaches that identified broad topics of increased relevance in these disease groups. Additionally, we applied rigorous hyperparameter search to identify the most parsimonious LDA solution, cognizant that the requirement of predefining topic number in LDA may lead to imperfect topic selection for a given corpus.

There are limitations to our work. As mentioned above, using word frequencies for inference assumes that each word has one and only one meaning. We also focused this word frequency analysis on social worker notes, without analyzing how these word frequencies vary across disease categories in other note types. On the topic modeling analyses, our method to ensure topic coherence required assigning topic labels to around 20 clusters for more than 10 note cohorts with 5 iterations each made the manual interpretation time-consuming and error-prone. Thus, we developed topic labeling heuristics that allow us to assign topics to the individual clusters (see **Methods**). We tested that this heuristic could perform as well as manual topic assignment, however, it still can be improved. In the future, it may be interesting to revisit our heuristic to expand upon the topic clusters further to make them more generalizable. Still, we believe this approach serves as an improvement over previous work in the field where human interpretation was the only method.

**CONCLUSION**

Social work notes contain rich and unique information about social determinants of health factors. Without using notes, it would be impossible to consider many factors in analyzing health outcomes. Latent dirichlet allocation topic modelling of social work notes is a scalable and unsupervised approach for characterizing the impact of social determinants of health on healthcare system and community public health. Although human inference and interpretation are necessary for topic modeling, several computation-based evaluation approaches were implemented to assist in discovering robust results. In addition, the findings indicate that different categories of notes emphasize on different aspects of social determinants of health. Understanding this information will help guide future work using clinical notes to study social determinants of health.

**ACKNOWLEDGEMENT**

We thank all researchers, clinicians, social workers who help collect clinical notes data in our UCSF Information Commons. We thank everyone in Dr. Atul J. Butte's lab for the helpful discussion. We thank everyone who has been helping to construct and organize the UCSF Information Commons. We thank Wynton high-performance computing (HPC) cluster for the computation capacity support.

**COMPETING INTERESTS**

AJB is a co-founder and consultant to Personalis and NuMedii; consultant to Mango Tree Corporation, and in the recent past, Samsung, 10x Genomics, Helix, Pathway Genomics, and Verinata (Illumina); has served on paid advisory panels or boards for Geisinger Health, Regenstrief Institute, Gerson Lehman Group, AlphaSights, Covance, Novartis, Genentech, and Merck, and Roche; is a shareholder in Personalis and NuMedii; is a minor shareholder in Apple, Meta (Facebook), Alphabet (Google), Microsoft, Amazon, Snap, 10x Genomics, Illumina, Regeneron, Sanofi, Pfizer, Royalty Pharma, Moderna, Sutro, Doximity, BioNtech, Invitae, Pacific Biosciences, Editas Medicine, Nuna Health, Assay Depot, and Vet24seven, and several other non-health related companies and mutual funds; and has received honoraria and travel reimbursement for invited talks from Johnson and Johnson, Roche, Genentech, Pfizer, Merck, Lilly, Takeda, Varian, Mars, Siemens, Optum, Abbott, Celgene, AstraZeneca, AbbVie, Westat, and many academic institutions, medical or disease specific foundations and associations, and health systems. AJB receives royalty payments through Stanford University, for several patents and other disclosures licensed to NuMedii and Personalis. AJB's research has been funded by NIH, Peraton (as the prime on an NIH contract), Genentech, Johnson and Johnson, FDA, Robert Wood Johnson Foundation, Leon Lowenstein Foundation, Intervalien Foundation, Priscilla Chan and Mark Zuckerberg, the Barbara and Gerson Bakar Foundation, and in the recent past, the March of Dimes, Juvenile Diabetes Research Foundation, California Governor's Office of Planning and Research, California Institute for Regenerative Medicine, L'Oreal, and Progenity. The authors have declared that no competing interests exist.

**MAIN FIGURES**

**FIGURE 1**

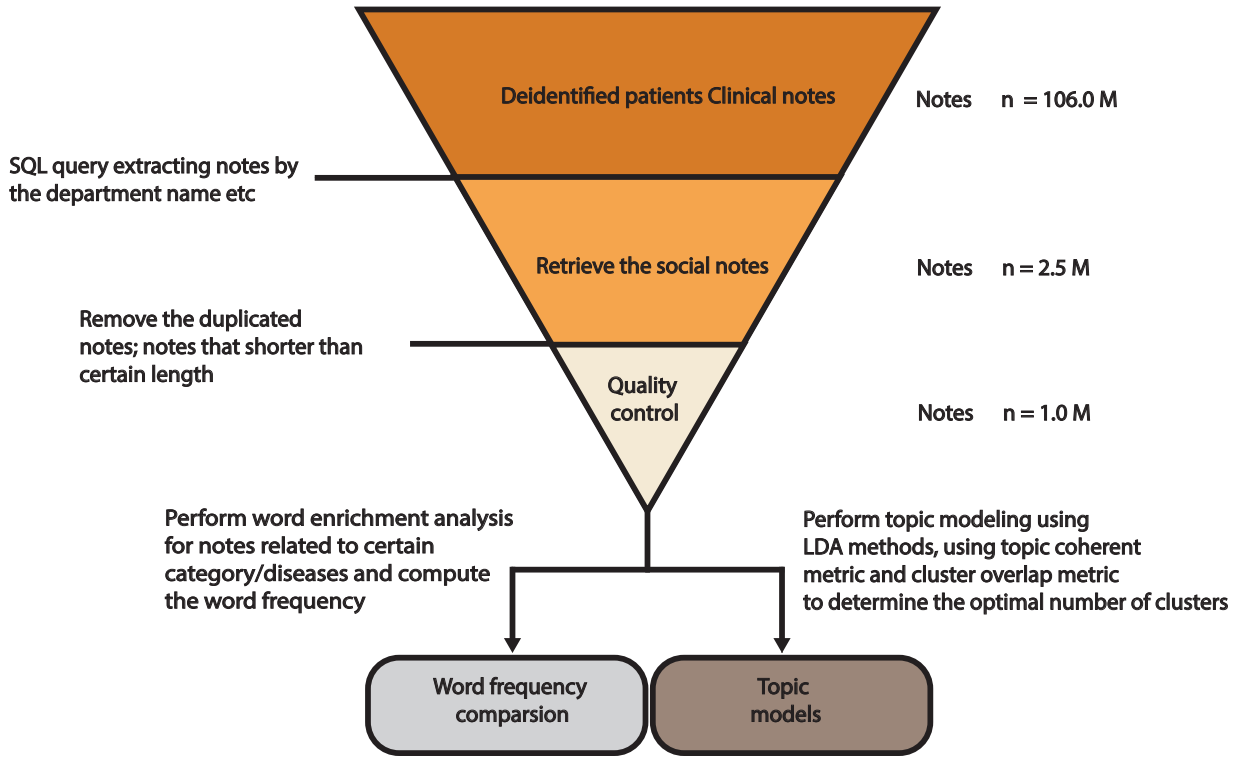

Cohort selection and notes analysis workflow

**Figure 1.** Retrieval of clinical social work notes for the study. The social work notes from the UCSF Information Commons between 2012 and 2021 were initially retrieved. Notes that were duplicated or extremely short were excluded, which resulted in a corpus of 0.95 million notes. Later, the notes were analyzed using two methods: word frequency calculation (Bottom Left) and topic modeling (Bottom Right). Later, the word frequency was compared between different disease chapters. For topic modeling, Latent Dirichlet Allocation was used to identify the topics in individual social work notes. Topic coherence metric and Jaccard distance were implemented to decide the optimal clustering results.

**FIGURE 2**

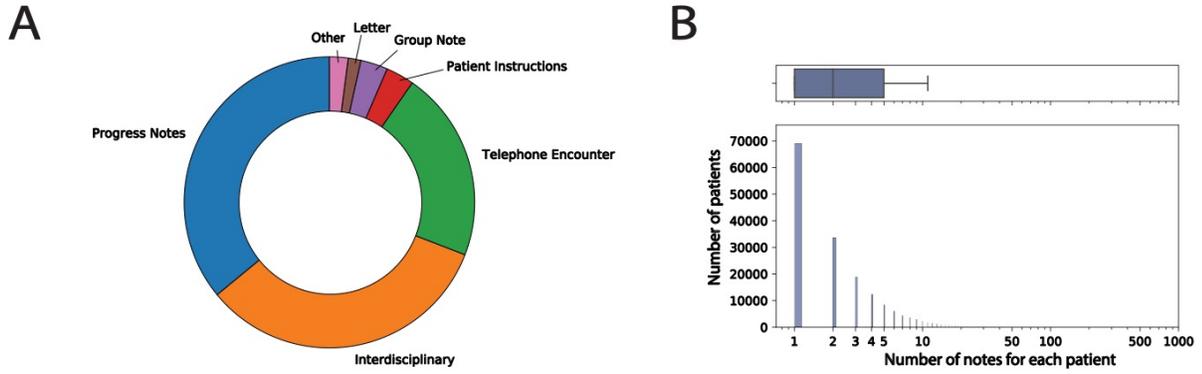

**Figure 2.** Data exploration on social work notes. A. Pie chart showing the proportions of patients in different categories. B. Boxplot and histogram showing the number of notes for the individual patients. The scale of x-axis is log10-transformed. The mode, mean, and median are 1, 5.8, and 2.

**FIGURE 3**

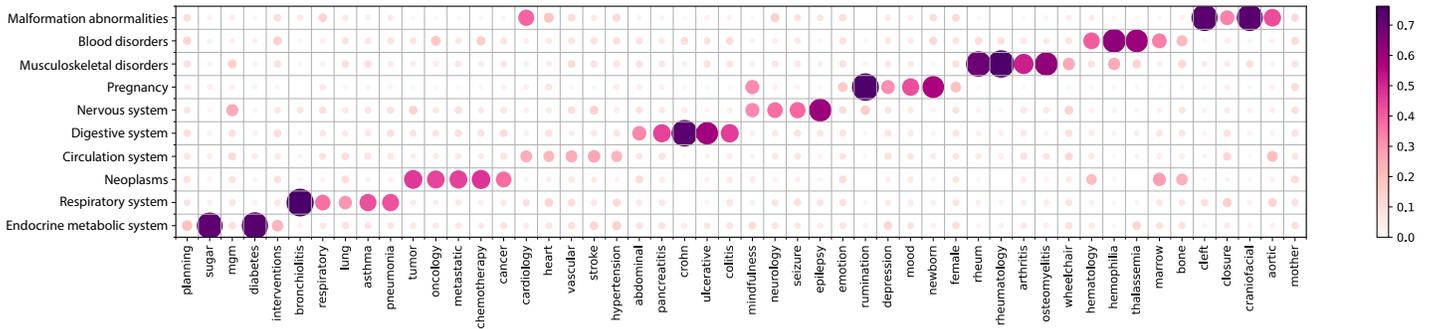

**Figure 3.** Word frequency calculation for social work notes associated with each ICD-10 chapter. The proportion of the words in social work notes associated with each ICD-10 chapter is shown by the heatmap.

**FIGURE 4**

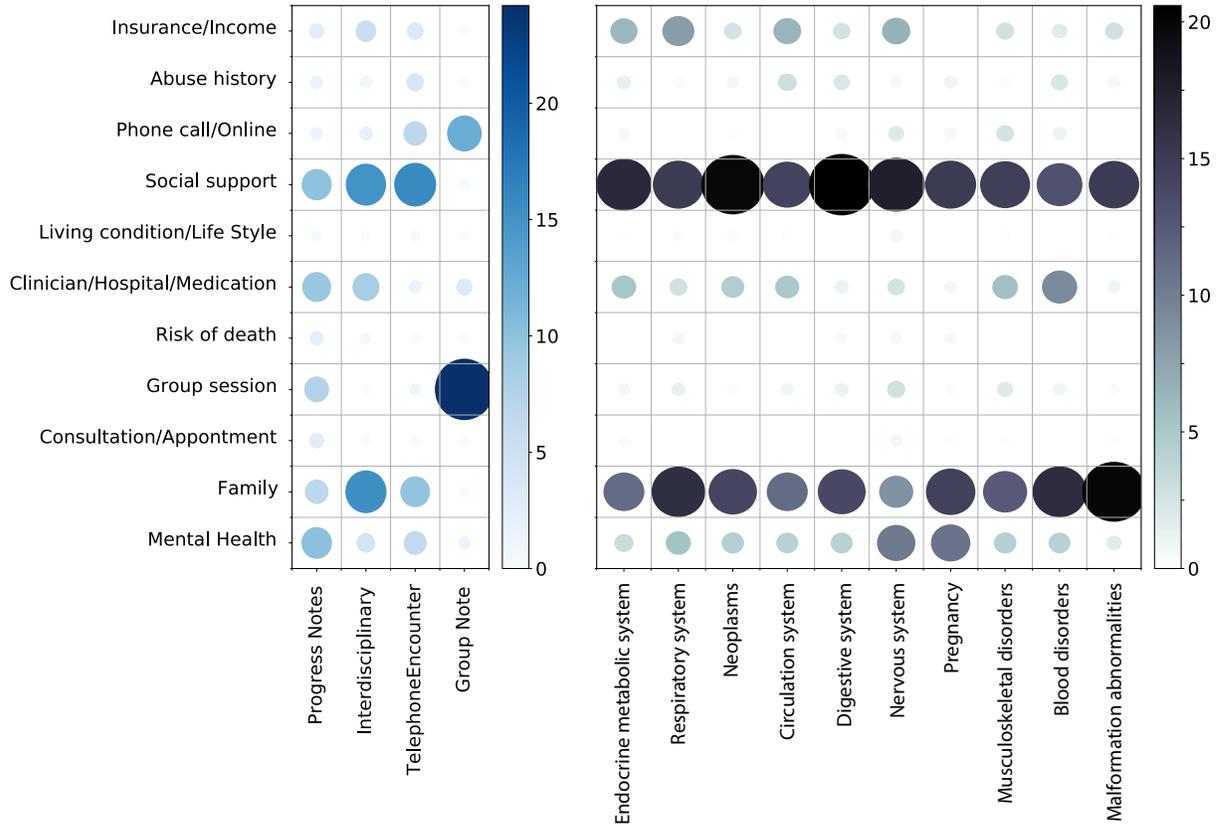

**Figure 4. Topic proportion comparison for different categories.** A. Topic proportion comparison for different note types. B. Topic proportion comparison for different disease chapters. Size and color of the circle represent proportion of each topic.

**Table 1.** Topic modeling results for all social work notes. Each row is an inferred topic, which is composed of 10 words.

| Clusters | Key words |
|---|---|
| 1 | goal, anxiety, problem, term, depression, mood, therapy, symptom, long, treatment |
| 2 | recommendation, wife, education, treatment, patient, form, appearance, ongoing, advocate, trauma |
| 3 | hospital, self, day, pain, other, connection, recent, feeling, side, number |
| 4 | mother, father, family, room, information, nurse, source, concrete, control, instruction |
| 5 | session, consultation, telehealth, location, time, tool, objective, parking, other, treatment |
| 6 | parent, family, school, child, sister, support, place, year, well, initial |
| 7 | group, intervention, patient, discussion, response, time, summary, progress, participant, skill |
| 8 | risk, chronic, thought, normal, imminent, status, testing, intervention, speech, suicide |
| 9 | client, health, service, caregiver, mental, therapist, therapy, behavioral, individual, group |
| 10 | well, when, time, week, also, able, state, more, friend, very |
| 11 | social, service, support, family, assessment, medical, time, note, concern, ongoing, |
| 12 | care, home, plan, phone, contact, work, information, resource, call, support |
| 13 | time, clinician, name, date, code, behavior, risk, number, plan, provider |
| 14 | history, child, other, factor, current, none, substance, abuse, psychiatric, year |
| 15 | donor, donation, potential, employment, understanding, risk, decision, independent, process, care |
| 16 | night, morning, hour, sleep, house, already, less, past, aggressive, evening |
| 17 | transplant, medication, post, support, health, insurance, husband, psychosocial, message, history |

SUPPLEMENTARY MATERIALS

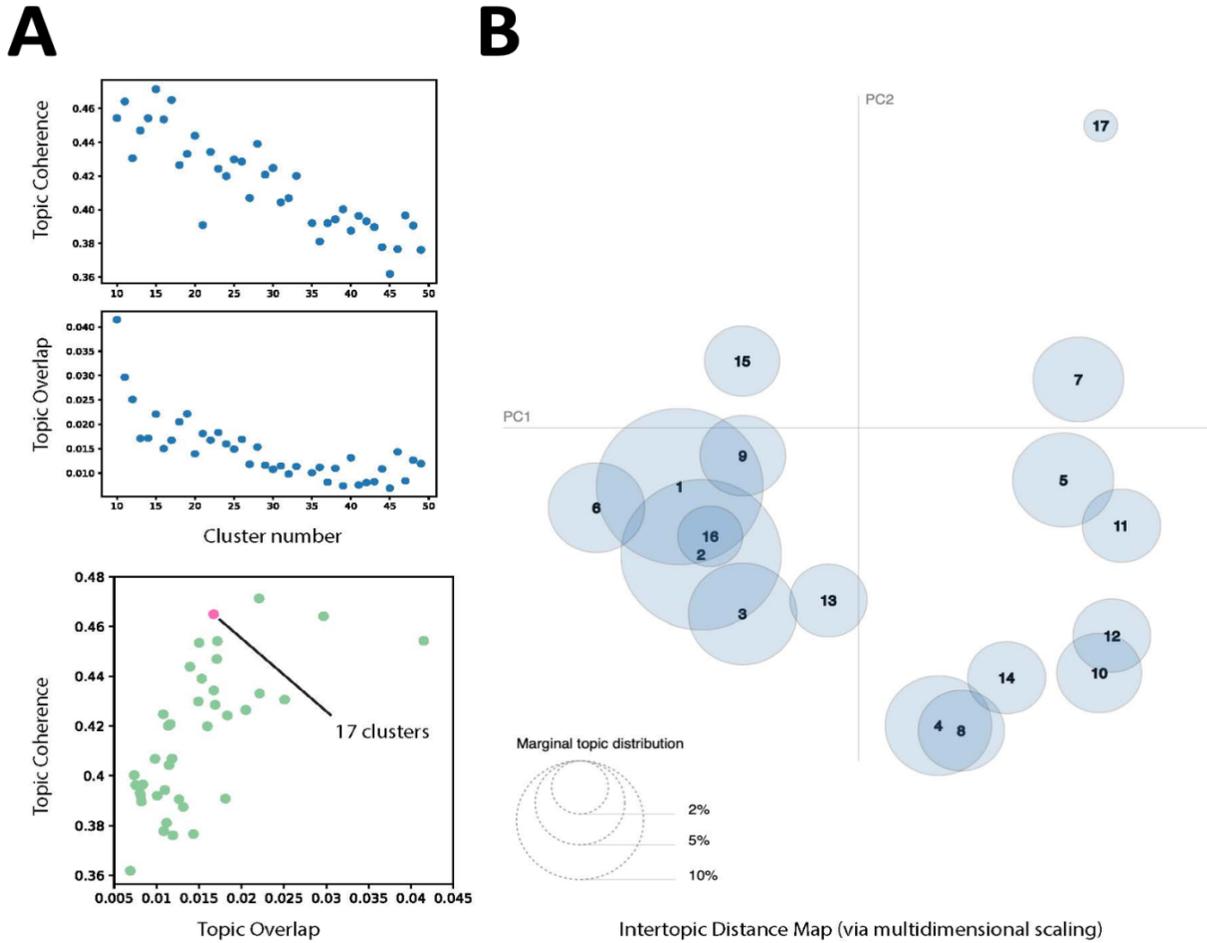

**S. Figure 1. Topic modeling clustering on whole social work notes.** A. Pipeline for determining the optimal number of clusters for the LDA method. The top: the plot of cluster number versus the topic coherence metric; The middle: the plot of the cluster number versus the topic overlap metric (measured by jaccard similarity metric); The bottom: the plot of the topic overlap metric versus the topic coherence metric. The number of clusters is chosen as 17 because it has the lowest topic overlap metric value while having the highest topic coherence metric value (see **Methods**) B. Inter-topic distance mapping for the individual cluster. Each circle represents an inferred topic. The coordinates for each circle correspond to the first two Principal components. The radius size indicates the frequency of topic existence on each note.

**S. Table 1:** Descriptive statistics for clinical social work notes corpus and contributing patient samples. Few Notes: Number of notes <= 1; Several Notes: 2<=Number of notes < 5; Many Notes: Number of notes >= 5.

|  | Few Notes (N=69211) | Several Notes (N=65100) | Many Notes (N=47333) | Overall (N=181644) |
|---|---|---|---|---|
| **Sex** | | | | |
| Female | 36372 (52.6%) | 34285 (52.7%) | 24730 (52.2%) | 95387 (52.5%) |
| Male | 32608 (47.1%) | 30626 (47.0%) | 22401 (47.3%) | 85635 (47.1%) |
| Unknown | 231 (0.3%) | 189 (0.3%) | 202 (0.4%) | 622 (0.3%) |
| **Ethnicity** | | | | |
| Hispanic/Latino | 14891 (21.5%) | 14451 (22.2%) | 12044 (25.4%) | 41386 (22.8%) |
| Not Hispanic/Latino | 48758 (70.4%) | 46011 (70.7%) | 33249 (70.2%) | 128018 (70.5%) |
| Unknown | 5562 (8.0%) | 4638 (7.1%) | 2040 (4.3%) | 12240 (6.7%) |
| **Race** | | | | |
| Asian | 8651 (12.5%) | 8578 (13.2%) | 5610 (11.9%) | 22839 (12.6%) |
| Black/African | 7153 (10.3%) | 7148 (11.0%) | 6819 (14.4%) | 21120 (11.6%) |
| Other | 17594 (25.4%) | 16922 (26.0%) | 13207 (27.9%) | 47723 (26.3%) |
| Unknown | 6683 (9.7%) | 5493 (8.4%) | 2637 (5.6%) | 14813 (8.2%) |
| White | 29130 (42.1%) | 26959 (41.4%) | 19060 (40.3%) | 75149 (41.4%) |
| **Age** | | | | |
| median (Q1-Q3) | 32 (11 - 58) | 35 (13 - 59) | 30 (10 - 57) | 33 (12 - 58) |
| Missing | 111 (0.2%) | 135 (0.2%) | 175 (0.4%) | 421 (0.2%) |

**S. Table 2.** Most frequent ICD-10 codes for patients with social work notes

| ICD10 code | Diagnosis | Notes | Patients |
|---|---|---|---|
| F01-F99 | Mental, Behavioral and Neurodevelopmental disorders | 106348 | 15388 |
| Z00-Z99 | Factors influencing health status and contact with health services | 36399 | 25995 |
| R00-R99 | Symptoms, signs and abnormal clinical and laboratory findings, not elsewhere classified | 25011 | 18820 |
| E00-E89 | Endocrine, nutritional and metabolic diseases | 13307 | 8488 |
| S00-T88 | Injury, poisoning and certain other consequences of external causes | 10508 | 5092 |
| I00-I99 | Diseases of the circulatory system | 9832 | 8450 |
| K00-K95 | Diseases of the digestive system | 7711 | 6267 |
| G00-G99 | Diseases of the nervous system | 7589 | 6187 |
| O00-O9A | Pregnancy, childbirth and the puerperium | 7370 | 5628 |
| D50-D89 | Diseases of the blood and blood-forming organs and certain disorders involving the immune mechanism | 6289 | 4598 |
| Q00-Q99 | Congenital malformations, deformations and chromosomal abnormalities | 6041 | 4507 |
| C00-D49 | Neoplasms | 6025 | 4515 |
| N00-N99 | Diseases of the genitourinary system | 5570 | 4926 |
| J00-J99 | Diseases of the respiratory system | 5294 | 4690 |
| M00-M99 | Diseases of the musculoskeletal system and connective tissue | 5018 | 4228 |
| P00-P96 | Certain conditions originating in the perinatal period | 4700 | 4426 |
| A00-B99 | Certain infectious and parasitic diseases | 2867 | 2536 |
| V00-Y99 | External causes of morbidity | 2104 | 2053 |
| L00-L99 | Diseases of the skin and subcutaneous tissue | 2023 | 1853 |
| H00-H59 | Diseases of the eye and adnexa | 889 | 841 |
| H60-H95 | Diseases of the ear and mastoid process | 562 | 516 |
| U00-U85 | Codes for special purposes | 85 | 84 |

**S. Table 3:** Topic assignment heuristic. The words in the *Keywords* column are the representative words used to define the topics.

| Topics | Keywords |
|---|---|
| Mental health | mental, depression, anxiety, mood, psychological, physical, cognitive, emotional, mind, psychiatric |
| Family | family, parent, father, mother, child, children, sister, parents, relatives, clan, childhood, friends |
| Consultation/Appointment | appointment, consultation, consult, questionnaire, question, advice, biographical, wikipedia, relevant, questions, know, documentation |
| Group session | group, intervention, session, interpers, community, class, organization, together, part, organization |
| Risk of death | suicide, suicidal, risk, crisis, homicide, murder, commit, bombing, murdered, murders, bomber, killing, convicted, victims |
| Clinician/Hospital/Medication | patient, medication, hospital, medical, clinic, clinician, treatment, therapy, surgery, symptoms, patients, drugs, diagnosis, treatments, prescribed |
| Living condition/Lifestyle | shelter, housing, house, living, sleep, bedtime, building, buildings, urban, employment, suburban, campus, acres |
| Social support | social, service, support, referral, recommendation, recommend, worker, resource, supports, provide, supporting, supported, allow, providing, assistance, benefit, help |
| TelephoneEcounter/Online communication | telehealth, phone, call, video, telephone, mobile, wireless, gsm, cellular, dial, email, calling, networks, calls, messages, telephones, internet |
| Abuse history | abuse, history, addiction, alcohol, drugs, allegations, victim, violence, sexual, rape, dependence |
| Insurance/Income | insurance, income, coverage, financial, contracts, banking, finance, liability, private, pay |